\pdfoutput=1

\documentclass[11pt]{article}

\usepackage{acl}

\usepackage{times}
\usepackage{latexsym}

\usepackage[T1]{fontenc}

\usepackage[utf8]{inputenc}

\usepackage{microtype}
\usepackage{makecell}
\usepackage{graphicx}
\usepackage{multirow}

\title{A Hybrid Model of Classification and Generation for Spatial Relation Extraction}

\author{Feng Wang \\
  fwangsakura \texttt{@stu.suda.edu.cn} \\\And
  Peifeng Li \\
  pfli \texttt{@suda.edu.cn} \\\And
  Qiaoming Zhu \\
  qmzhu \texttt{@suda.edu.cn} \\
  }

\begin{document}
\maketitle
\begin{abstract}
Extracting spatial relations from texts is a fundamental task for natural language understanding and previous studies only regard it as a classification task, ignoring those spatial relations with  null roles due to their poor information. To address the above issue, we first view spatial relation extraction as a generation task and propose a novel hybrid model HMCGR for this task. HMCGR contains a generation and a classification model, while the former can generate those null-role relations and the latter can extract those non-null-role relations to complement each other. Moreover, a reflexivity evaluation mechanism is applied to further improve the accuracy based on the reflexivity principle of spatial relation. Experimental results on SpaceEval show that HMCGR outperforms the SOTA baselines significantly.
\end{abstract}


\section{Introduction}
Spatial relation extraction focuses on identifying the relationship between two geographical entities in natural language texts. Currently, only a few studies focused on this task in the NLP community, while most studies aimed at the other tasks of relation extraction, such as temporal and causal relation extraction. However, spatial information is one kind of critical information for natural language understanding, which can benefit the downstream NLP applications, such as spatial domain query \cite{zhang2020joint}, spatial reference \cite{yang2020robust} and data forecasting \cite{song2020spatial}.

\begin{figure}[htp]
    \centering
    \includegraphics[width=0.5\textwidth]{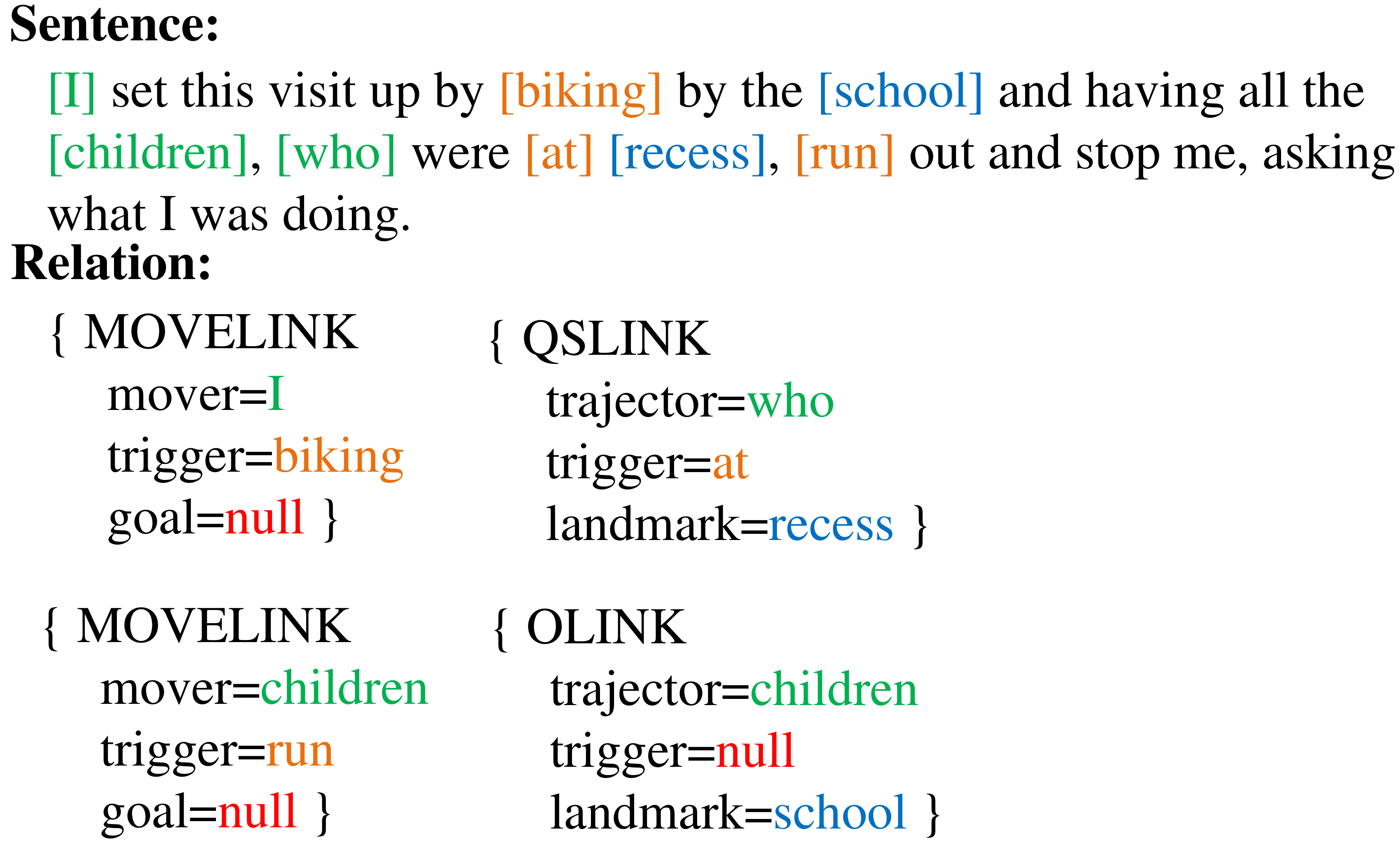}
    \caption{ An example in SpaceEval with null-role and non-null-role.}
    \label{fig:galaxy}
    \vspace{-0.4cm}
\end{figure}

Various kinds of schemes have been proposed to
represent spatial relation. As one of the SemEval evaluation tasks, SpaceEval \cite{DBLP:conf/semeval/PustejovskyKMLD15} proposes an annotation scheme adopted from ISO-space  \cite{pustejovsky2011using} and its goals include identifying and classifying items from an inventory of spatial concepts, such as topological relations, orientational relations, and motion, etc. Commonly, this task needs to extract the spatial elements and classify static and dynamic spatial relations into three types: the move link (MOVELINK), the qualitative spatial link (QSLINK),  and the orientation link (OLINK). MOVELINK connects motion-events with corresponding mover-participants as a triplet of three roles \emph{(mover, goal, trigger)}, while QSLINK and OLINK refer to the topological relation and non-topological relation between two spatial elements, respectively, and are formalized as a triplet  of three roles \emph{(trajector, landmark, trigger)}.

Following previous work, we also simplified the whole task as Figure \ref{fig:galaxy} and only focus on extracting QSLINK/OLINK/MOVELINK from texts. Thus, a spatial relation is defined as a triplet with three spatial types. 
The spatial relation can be divided into two classes: null-role and non-null-role relations. The former refers to a relation containing null-value roles, such as two MOVELINKs and one OLINK in Figure \ref{fig:galaxy}, while the latter (e.g., QSLINK in Figure \ref{fig:galaxy}) refers to a relation whose three roles are fulfilled the values extracted from sentences.

Almost all previous studies regarded spatial relation extraction as a classification task using traditional machine learning \cite{DBLP:conf/semeval/NicholsB15, DBLP:conf/emnlp/DSouzaN15} or neural network methods \cite{DBLP:conf/ecir/RamrakhiyaniPV19,shin-etal-2020-bert}.  Those classification models work well on extracting those non-null-role relations due to their rich information. However, they often suffer from those null-role relations. The reason is that some information is missing in these relations. 
Moreover, they also cannot benefit from the knowledge of the spatial schema, such as the roles and their relations. 

In the annotation stage, annotators usually not only annotate relations and relation types, but also provide a description or basis for their annotation implicitly. Therefore, we hope the model can simulate a human and provide a target sentence instead of a simple label index for understanding the spatial relation deeper. The target sentence in generation models can describe the relation between all spatial elements and it allows null slots (i.e., roles) to exist. 
Thus, the generation model not only can more explicitly learn the semantics of the spatial relations through such a form of the learning goal, but also can generate those null-role relations.

Moreover, the classification model and the generation model have their complementary advantages. The former usually has better performance on no-null-role relations, while the latter can introduce prior knowledge to capture the semantics of  null-role relations better and its results are in a natural language expression with stronger interpretability \cite{jiang2021not}. Therefore, we combine the advantages of the classification and generation models to further capture different knowledge.

In this paper, we propose a novel hybrid model \textbf{HMCGR} (\textbf{H}ybrid \textbf{M}odel of \textbf{C}lassification, \textbf{G}eneration and \textbf{R}eflexivity) for spatial relation extraction, which contains a generation model and a classification model. Specially, the former can generate those null-role relations and the latter can extract those non-null-role relations to complement each other. Moreover, a reflexivity evaluation mechanism is applied to further improve the accuracy based on the reflexivity principle of spatial relation. Experimental results on the SpaceEval dataset shows that our HMCGR outperforms the SOTA baselines significantly.

\section{Related Work}
Various kinds of schemes have been proposed to represent spatial relations. SpatialML \cite{mani2010spatialml} characterized directional and topological relations among locations in terms of a region calculus. The SpRL task \cite{2011Spatial} developed a semantic role labeling scheme focusing on the main roles in spatial relations. Spatial relation extraction was introduced as subtask at SemEval 2012 \cite{kordjamshidi2012semeval}, SemEval 2013 \cite{kolomiyets2013semeval} and  SemEval 2015 \cite{DBLP:conf/semeval/PustejovskyKMLD15}. As the Task 8  of SemEval 2015, SpaceEval proposed an annotation scheme adopted from ISO-space,  and it enriched SpRL’s semantics by refining the granularity. Most of previous studies were evaluated on this dataset. 

The task of spatial relation extraction can be divided into traditional machine learning and neural network methods. The former highly relies on manual features or explicit syntactic structures. \citet{DBLP:conf/semeval/NicholsB15} used a CRF layer to extract spatial elements, and then introduced SVM to classify  spatial relations. \citet{DBLP:conf/emnlp/DSouzaN15} proposed a Sieve-based model where various kinds of manual features are generated by a greedy feature selection technique. \citet{DBLP:conf/semeval/SalaberriAZ15} introduced external knowledge as a supplement to spatial information, in which WordNet and PropBank provided information on many spatial elements. \citet{Kim2016ExtractingSE} proposed a Korean spatial relation extraction model using dependency relations to find the proper elements to fulfill roles. 

With the wide application of neural network, \citet{DBLP:conf/ecir/RamrakhiyaniPV19} generated candidate relations by dependency parsing and classified the candidates with a BiLSTM model.  \citet{shin-etal-2020-bert} first used BERT-CRF to extract the spatial roles and then introduced R-BERT \cite{wu2019enriching} to extract the spatial relations. Besides, a few studies focused on multi-modal spatial relation extraction. For example, \citet{DBLP:journals/corr/abs-2007-09551} proposed a spatial BERT which gives two spatial entities and a picture to determine spatial relations. 


\begin{figure*}[htp]
    \centering
    \includegraphics[width=1.0\textwidth,height=0.7\textwidth]{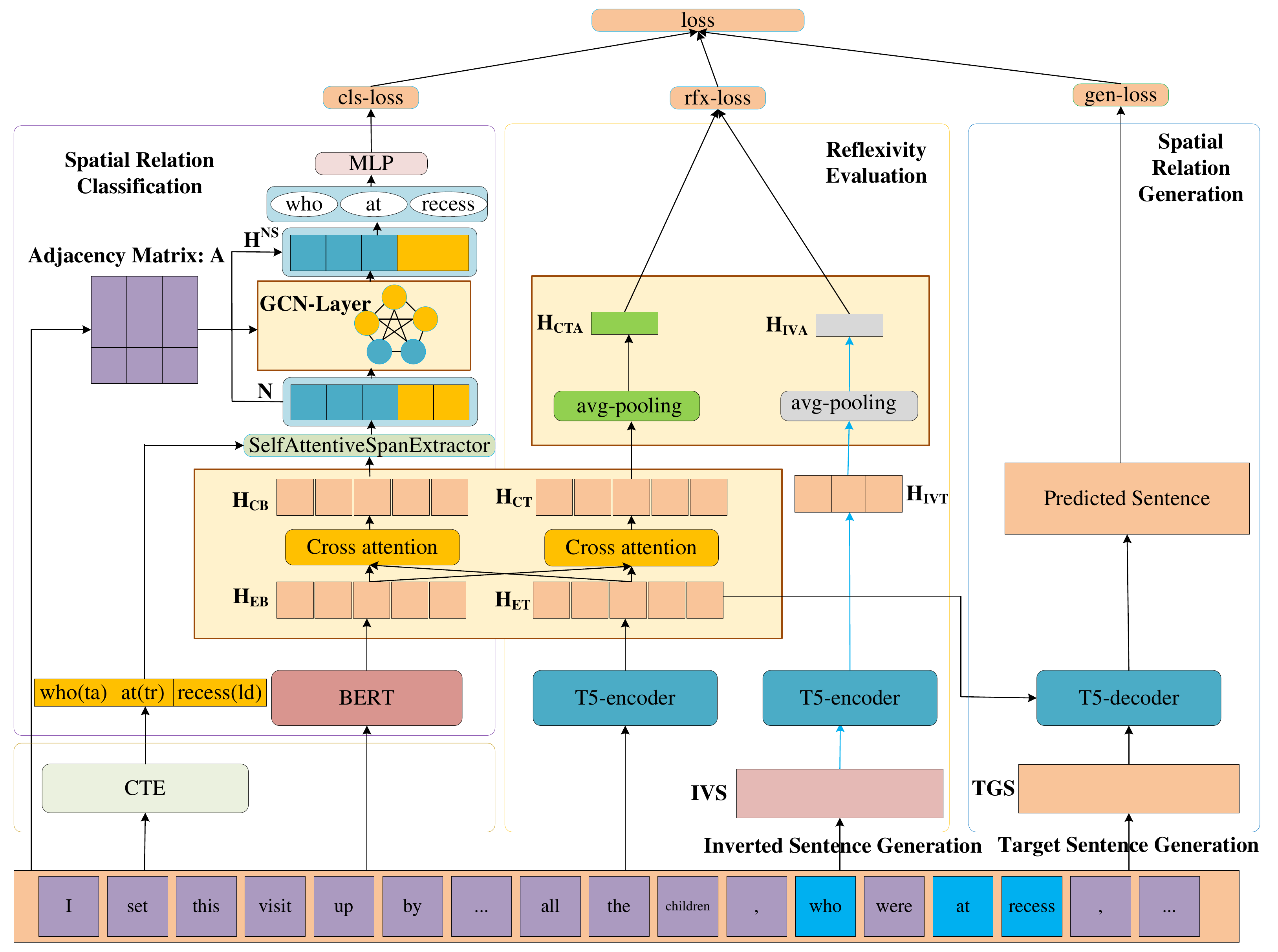}
    \caption{Overall structure of our HMCGR.}
    \label{fig:B}
\end{figure*}

\section{HMCGR}

 Figure \ref{fig:B} shows the overall architecture of our model HMCGR. As a whole, HMCGR can be divided into four modules, i.e., candidate triplet extraction (CTE), spatial relation classification (CLS), spatial relation generation (GEN), and Reflexivity evaluation (RFX). 
 
 The module CTE is first used to extract spatial elements and spatial roles from a raw sentence to obtain candidate triplets by a BERT-CRF model. And then the candidate triplets and the raw sentence are fed to the module CLS, which uses a BERT encoder and a T5 encoder to encode the sentence, respectively, and apply a  GCN (Graph Convolutional Networks) layer to capture the sentence structure. Simultaneously, the module GEN uses a T5 decoder to generate a target sentence following a specific template, and the module RFX uses the cosine function to calculate the similarity between the original sentence and its inverted sentence to further improve the accuracy. 

\subsection{CTE: Candidate Triplet Extraction}
\begin{figure}[htp]
    \centering
    \includegraphics[width=0.5\textwidth]{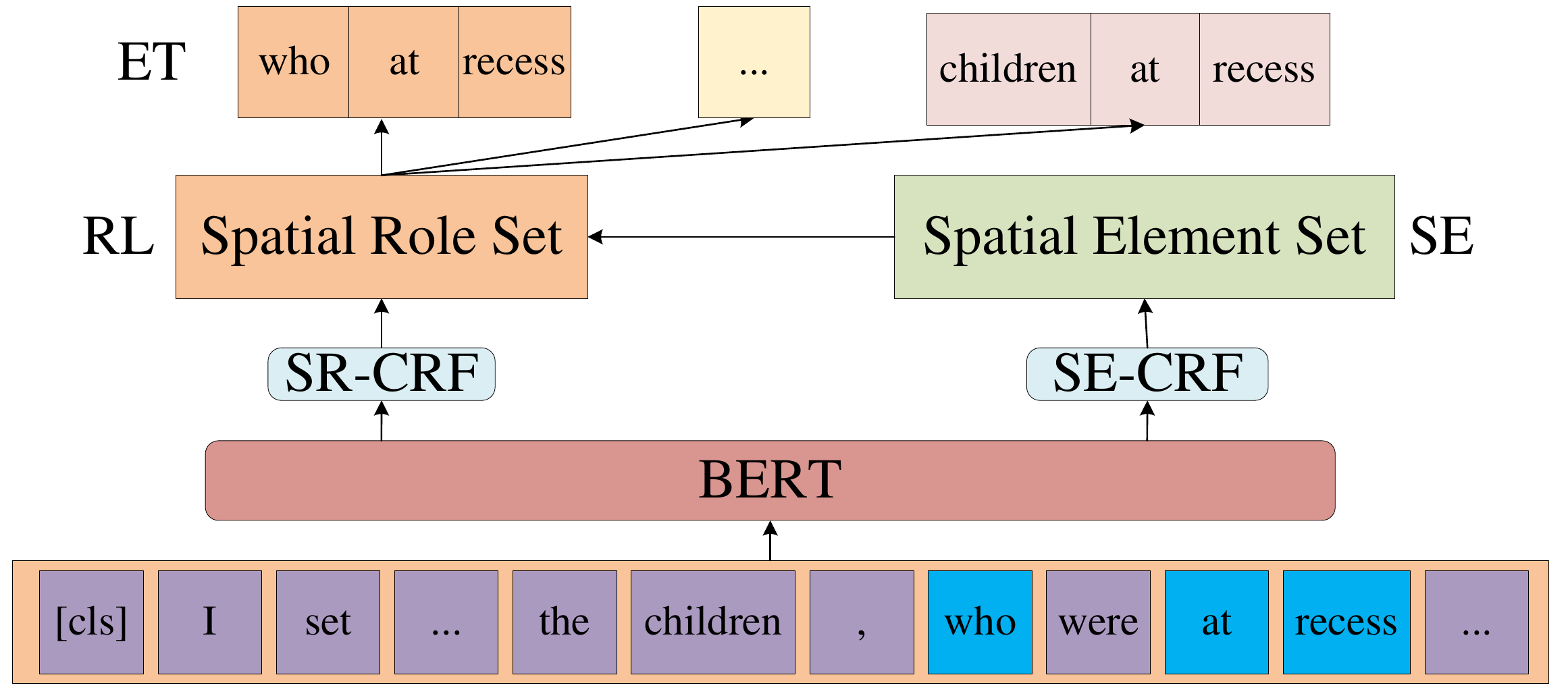}
    \caption{Overview of candidate triplet  extraction.}
    \label{fig:C}
\end{figure}

Since a spatial relation is represented as a triplet with its relation type MOVELINK, OLINK or QSLINK, the first step of HMCGR is to extract candidate triplets from raw texts as much as possible. Similar to \citet{shin-etal-2020-bert}, we also use the BERT+CRF model for spatial role extraction, as showed in Figure \ref{fig:C}. Spatial role extraction is a task to form candidate triplets, which extracts the spatial elements from texts and then assigns a role to each extracted element.

Formally, the input is a token sequence $X=(x_{1},...,x_i,...,x_n)$ where $x_i$ is the $i$-th token in a sentence $S$. We feed $X$ with the label CLS to BERT to obtain a new embedding $H_B$ from BERT which $H_B=[b_{1},...,b_i,...,b_n]$ where $b_{i} \in R^{d_{b}}$ and $R^{d_{b}}$ is the pre-defined spatial role set. 

In Figure \ref{fig:C}, there are two CRF layers with the input embedding $H_B$, i.e., the Spatial Element CRF SE-CRF  and the Spatial Role CRF SR-CRF. We use SE-CRF to obtain  the spatial element set $SE=[se_1,...,se_i,...,se_m]$  in $S$ where $se_i$ is a spatial element, and use SR-CRF to obtain the role set $RL = [rl_1,...,rl_i,...,rl_m]$ for all elements where $rl_i$ is the spatial role of the element $se_i$.

 We simply apply a multi-task framework to train these two CRFs and they share the same BERT encoder layer. Take the sentence in Figure \ref{fig:galaxy} as example, we can extract six spatial elements ``children'',  ``school'',  ``in'', ``who'', ``at'' and ``recess'', whose roles are \emph{Spatial Entity}, \emph{Place}, \emph{Spatial Signal}, \emph{Spatial Entity},  \emph{Spatial Signal} and \emph{Place}, respectively.

Due to CTE is the first stage of HMCGR, we tend to generate all possible spatial role triplets for the subsequent CLS module to achieve high recall. Hence, we first split  the set $SE$ into three subsets : 1) $TM$=\{$Trajecto$r, $Mover$\}, 2) $LG$=\{$Landmark$, $Goal$\}, and 3) $TR$=\{$Trigger$\} according to their roles. Take the above elements for example, ``children'' and ``who'' belong to $TM$, while ``school'' and ``recess'' belong to $LG$ and the others belong to $TR$.

\vspace{-0.2cm}
 
 Finally, we enumerate possible triplets as candidates following the spatial relation definition. Commonly, some triplets may have the roles with null values, as the role $trigger$ showed in Figure \ref{fig:galaxy}, because its element does not mention in the according sentence. If we enumerate all possible triplets including null roles as candidates, this will introduce enormous negative triplets into the candidate set and then harm the precision badly. For example, there are 27 ($3^3$) candidate triplets in the sentence in Figure \ref{fig:galaxy}, while only 4 are annotated triplets. Hence, we do not generate the triplets with null-value roles in the module CTE and the extracted candidate triplet set $ET=|TM|*|TR|*|LG|$ of the example in Figure \ref{fig:galaxy} is  as follows:  (who, at, recess), (who, in, school), (who, at, school), (who, in, recess), 
 (children, in, school), (children, at, school), 
 (children, in, recess) and (children, at, recess). 
 
\subsection{CLS: Spatial Relation Classification}
Following previous work, CLS is to classify the candidate triplets into four types, i.e., MOVELINK, OLINK, QSLINK, and null. If a triplet belongs to the type null, this triplet is a pseudo spatial relation. The reason that we introduce the type null to CLS is that there are lots of pseudo triplets extracted by CTE and they will harm the precision. 

\subsubsection{Encoding}

First, we simply use BERT and T5 to encode the sequence $X$\footnote{We add [cls] to the start of $X$ to obtain the sentence representation of BERT.} of the sentence $S$ to obtain the embeddings $H_{EB}=\{eb_{1},...,eb_{i},...,eb_{n}\}$ and $H_{ET}=\{et_{1},...,et_{i},...,et_{k}\}$, respectively. To make better use of the advantages of both two pre-training models, we use cross attention to represent the hidden layer state as follows. In this way, we can get the new  embeddings $H_{CB}=\{cb_{1},...,cb_{i},...,cb_{n}\}$ and  $H_{CT}=\{ct_{1},...,ct_{i},...,ct_{k}\}$ while the latter is used in the RFX module. 

\begin{equation}
cb_{i}=cross\_attention(eb_i, et_j)
\end{equation}
\begin{equation}
ct_{i}=cross\_attention(et_j, eb_i)
\end{equation}

Second, we incorporate the candidate triplets extracted by CTE into the above embedding $H_{cb}$ to enhance the representation of spatial elements. Specially, we introduce the SelfAttentiveSpanExtractor in AllenNLP to obtain the latent representation of three spatial roles $H_{tm}$, $H_{lg}$ and $H_{tr}$ as follows.
\begin{equation}
H_{y}=\sum_{i=y_{start}}^{y_{end}} W_{y_{i}} cb_{i}\\
\end{equation}
\noindent where $y \in \{tm, lg, tr \}$. $y_{start}$ and $y_{end}$ represent the start  and end position of a spatial element, respectively, and $W_{y_{i}}$ are learnable parameters. Besides, since BERT maybe splits a word into multiple word-pieces, we also use SelfAttentiveSpanExtractor to obtain word-level representation.

\subsubsection{Spatial GCN}

Most previous work ignored the function of demonstrative pronouns in spatial relation extraction. However, those pronouns can participate in various spatial relations. Inspired by \citet{DBLP:conf/naacl/PhuN21} in casual relation extraction, to capture the relationship between sentences and spatial roles, and make better use of sentence structure and anaphora, we introduce a spatial graph $G=\{N,E\}$ to CLS, where the node set $N=X \cup SE$, which are defined in subsection 3.1. We initialize four adjacency matrices ($A^{B}$, $A^{E}$, $A^{C}$, $A^{D}$) to represent four edge types in our graph $G$ as follows.

\emph{Sentence Boundary Edge}: Intuitively, relevant contextual information between the spatial elements within a sentence is helpful for this task. Hence, we create an undirected edge between two nodes if they are in the same sentence. Formally, we set $A^{B}_{i,j}=A^{B}_{j,i}=1$ if the nodes $n_i$ and $n_j$ ($n_i, n_j \in N$) in the same sentence; otherwise, 0.

\emph{Spatial Element Edge}: The intersections between the spatial elements and their containing tokens maybe share some useful information. Therefore, we create a spatial element edge between the spatial element and its token. Formally, we set $A^{E}_{i,j}=1$ if  $n_i$ contains $n_j$; otherwise, 0.

\emph{Coreference Edge}: According to our statistics, about 20\% of the spatial relations in SpaceEval are participated by demonstrative pronouns. Hence, we construct an edge from two nodes if one can reference the other. Formally, we set $A^{C}_{i,j}=1$ if  $n_i$ and $n_j$ are coreferential; otherwise, 0.

\emph{Dependency Edge}: Following previous work in NLP, we also create an edge if two nodes have the same parent node in the dependency tree. Formally, we set $A^{D}_{i,j}=A^{D}_{i,j}=1$ if $n_i$ and $n_j$ has the same parent node in a dependency tree. Besides, we utilize SpaCy \footnote{https://spacy.io/} to extract the dependency trees and coreference chains.


Due to the importance of different type edges, we conduct four learnable weight matrices $W^{y}$ ($y \in \{B, E, C, D\}$) to merge four type edges by their weights to an adjacency matrix $A$ as follows.

\begin{equation}
A_{i,j}=\sum_{y \in \{B, E, C, D\}} W^y_{i,j}  A^{y}_{i,j}
\end{equation}

Finally, we can easily construct the graph $G$ and formulate GCN for spatial information fusion to obtain its representation $H^{NS}$ as follows.
\begin{equation}
H^{NS} = GCN(G,N)
\end{equation}

\subsubsection{Classification}

By recording the node identifier of the spatial role in the currently processed triplet, we can get the latent representation of the spatial role in $NS$. Inspired by the idea of ResNet \cite{he2016deep}, we concat the BERT hidden state $H_y$ ( $y \in \{tm, lg, tr \}$) and the representation of the GCN nodes $H_y^{NS}$ as the final feature of the spatial roles as follows.
\begin{equation}
H^{'}_{y}=[H_{y};H_{y}^{NS}]
\end{equation}
\noindent where $H_{y}^{NS}$ represents the latent representation of the spatial roles in $H^{NS}$. Finally, a multi-layer perceptron  (MLP) is to classify the spatial relations and we calculate the cross-entropy loss as follows. 
\begin{equation}
y_{rel} = MLP([H^{'}_{tm};H^{'}_{tr};H^{'}_{lg}])
\end{equation}
\begin{equation}
L_{cls}=-\sum_{(tm,tr,lg) \in ET}log P(rel|tm,tr,lg)
\end{equation}
\noindent where $ET$ is the triplet set mentioned in subsection 3.1 and $rel$ is the relation of the triplet. 

\subsection{GEN: Spatial Relation Generation}
To reduce negative triplets in the CTE module, we only enumerate candidate triplets without null roles. This strategy can help CLS improve its precision. However, it also cannot extract those null-role relations. Our statistics on the SpaceEval dataset show 20\% of the annotated spatial relations have a null role. Hence, how to extract those null-role relations still is a challenge. To address this issue, we introduce a spatial relation generation module GEN to extract those null-role relations. Hence, HMCGR contains a classification and a generation model, and they can complement each other to address their shortcomings.


We introduce the pre-trained generation model T5 to our GEN, due to its excellent performance on many NLP applications \cite{DBLP:journals/jmlr/RaffelSRLNMZLL20}. Normally, there are two T5-decoding methods that can be used in our task, i.e., triplet or a normal sentence. In our experiments, we found that a structure normal sentence is suitable for the target sentence of T5, which contains the following three parts.

\emph{Referential Phrase Prefix}: To better use the coreference relation, we add a phrase with referential meaning to the target sentence and put this phrase in the beginning of the target sentence to let our GEN use this useful information.

\emph{Relation Name}: To get the type of spatial relation, we designed a slot of spatial relation name for the target sentence.

\emph{Relation Explanation}: To decode spatial relations more quickly and conveniently, we design a generate structured sentence with $<$pad$>$ spatial role slots as our target sentence.

Specifically, the form of target sentence is as follows: ``The token ``$pronoun$'' stands for ``$noun$'', and $<pad>$ $qslink$ $<pad>$ can be describe as following : the first element is $<pad>$ $tm$ $<pad>$, the trigger is $<pad>$ $tr$ $<pad>$,  and the second element is $<pad>$ $lg$ $<pad>$.'' Take the candidate triplet (who, at, recess) as an example, we generate the following target statement $TGS$ for T5:``The token ``who'' stands for ``children'', and $<pad>$ qslink $<pad>$ can be describe as following : the first element is $<pad>$ who $<pad>$, the  trigger is $<pad>$ at $<pad>$, and the second element is $<pad>$ recess $<pad>$.''.

We feed a sentence representation $H_{ET}$ into T5-decoder and obtain a target sentence following the format of $TGS$, which can be translated into the form $Relation(tm, tr, lg)$. It is worth noting that one of $tm$, $tr$ and $lg$ may be null, and we can obtain those null-role relations. Finally, T5 generates a token or phrase for each output position using softmax and then we can get the target sentence and the cross-entropy loss $L_{gen}$ is defined as follows. 
\begin{equation}
L_{gen}=T5_{decoder}(TGS,H_{ET})
\end{equation}

\subsection{RFX: Reflexivity Evaluation}
Our CLS and GEN can extract spatial relations from different perspectives and complement each other effectively. However, the performance of GEN is still lower than CLS, because it suffers from the limited training data and the high rate of negative and positive instances in this task.

Most spatial relations have the attribute of reflexivity due to their nature. For example, "A in B" equals to "B out of A" in spatial relation. According to the reflexivity of spatial relation, we design a similarity-based reflexivity evaluation mechanism to help GEN improve its performance. RFX first creates an inverted sentence $IVS$ according to the original sentence $S$ and a candidate triplet $et$, and then uses the cosine function to calculate the similarity of their embeddings. If two sentences are similar, the candidate triplet $et$ will be regarded as a spatial relation with high probability.

For a sentence $S$ and a candidate triplet $et=(tm, tr,lg)$,  we first exchange the positions of two participants $tm$ and $lg$  in $S$, and then replace $tr$ with its antonym from an antonym dictionary. If $tr$ has more than one antonym, we randomly select one. The original sentence $S$ and the inverted sentence $IVS$ are fed to a T5-encoder to obtain the embeddings $H_{CT}$ using cross attention and $H_{IVT}$, respectively. The avg-polling is applied to the above two embeddings to capture their global features as follows.
\begin{equation}
H_{CTA}=avgpooling(H_{CT})
\end{equation}
\begin{equation}
H_{IVA}=avgpooling(H_{IVT})
\end{equation}

Finally, we design the spatial semantic loss $rfx-loss$ using a cosine similarity as follows.
\begin{equation}
L_{rfx}=1-cos(H_{CTA},H_{IVA})
\end{equation}

\subsection{Joint Training and Decoding}
In the training step, we train the classification model CLS and the generation model GEN together. To sum up, the overall loss $L$ of our model HMCGR consists of three parts as follows.

\begin{equation}
L=L_{cls}+L_{rfx}+L_{gen}
\end{equation}

Finally, the spatial relations are extracted by two models, i.e., CLS and GEN. The final spatial relation set is the union of their results. Besides, the module RFX is an effective auxiliary task to help GEN improve its performance.

\begin{table}
\centering
\begin{tabular}{lc}
\hline \textbf{Tool/Parameter} & \textbf{Version/Value}\\
\hline
Pytorch  & 1.7.0+cu110\\
Spacy  & 2.1.0\\
Allennlp  & 2.6.0\\
dgl-cu110  & 0.6.1\\
Learning rate  & 2e-5\\
Batch size  & 4\\
Random seed  & 1024\\
Hidden size of pre-training model  & 768\\
Optimizer  & AdamW\\
\hline
\end{tabular}
\caption{\label{font-table} Key parameters and tools used in our model. }
\end{table}

\begin{table}
\centering
\begin{tabular}{llll}
\hline
Model & P & R & F1\\
\hline
BERT+CRF	& 88.1	& 91.2	& 89.1\\
\hline
\end{tabular}
\caption{\label{roleresults}
The results of spatial role extraction.
}
\end{table}

\begin{table*}
\setlength\tabcolsep{5pt}
\centering
\begin{tabular}{cccc|ccc|ccc|ccc}
\hline
\multirow{2}{*}{Model}& 
\multicolumn{3}{c}{QSLINK}&               
\multicolumn{3}{c}{OLINK}& 
\multicolumn{3}{c}{MOVELINK}&          
\multicolumn{3}{c}{Overall}\\
\cline{2-4}  \cline{5-7} \cline{8-10} \cline{11-13}
& P & R & F1
& P & R & F1
& P & R & F1
& P & R & F1\\
\hline
{\shortstack{Sieve-Based}}
& {12.9} & {28.3} & {17.8}
& {\textbf{100}} & {31.2} & {47.5}
& {24.5} & {56.2} & {34.2}
& {45.8} & {38.5} & {41.8}\\

{\shortstack{WordNet}}
& {-} & {-} & {-}
& {-} & {-} & {-}
& {-} & {-} & {-}
& {54.0} & {51.0} & {53.0}\\

{\shortstack{SpRL-CWW }}
& {\textbf{66.1}} & {53.8} & {59.4}
& {69.1} & {51.7} & {59.1}
& {57.1} & {45.1} & {50.4}
& {63.6} & {50.1} & {56.1}\\

{\shortstack{BERT-base}}
& {\underline{45.1}} & {\underline{58.3}} & {\underline{50.5}}
& {\underline{71.0}} & {\underline{69.6}} & {\underline{70.2}}
& {\underline{62.7}} & {\underline{61.5}} & {\underline{62.1}}
& {62.7} & {59.8} & {61.2}\\

{\shortstack{HMCGR}}
& {53.5} & {\textbf{73.1}} & {\textbf{61.1}}
& {73.1} & {\textbf{85.2}} & {\textbf{78.6}}
& {\textbf{66.8}} & {\textbf{83.0}} & {\textbf{73.9}}
& {\textbf{64.3}} & {\textbf{79.2}} & {\textbf{70.9}}\\ 
\hline
\end{tabular}
\caption{\label{mainresults}
Performance comparison between the baselines and HMCGR on spatial relation extraction. Since BERT-base did not report the results on each category, we run their model to obtain the results (underlined).
}
\end{table*}

\section{Experimentation}

\subsection{Experimental Settings}
We evaluate our model on the latest dataset SpaceEval. According to the
official statistics, there are 1110 QSLINKs, 974 MOVELINKs and 287 OLINKs. We use the standard training/development/test set following previous work \cite{shin-etal-2020-bert} where the rate of the training set and the test set is 8:2. As for evaluation, we report Precision (P), Recall (R), and Micro-F1 score. We use Pytorch and Huggingface as our base tools and use the base versions of BERT and T5. The specific tool versions and key hyper-parameters are shown in Table \ref{font-table}.

Currently, only a few work focused on spatial relation extraction. To evaluate the effectiveness of our HMCGR, we conduct the following strong baselines for comparison: 1) \textbf{Sieve-Based} \cite{DBLP:conf/emnlp/DSouzaN15}, which used the sieve mechanism and syntactic parse trees to enhance the features in spatial relations; 2) \textbf{WordNet} \cite{DBLP:conf/semeval/SalaberriAZ15}, which used WordNet as an external knowledge to assist their task; 3) \textbf{SpRL-CWW} \cite{DBLP:conf/semeval/NicholsB15}, which is the SOTA traditional model using SVM and CRF classifiers  on the  GloVe features to extract the spatial relations; 4) \textbf{BERT-base} \cite{shin-etal-2020-bert}, which is the SOTA neural network model using a BERT-based neural network model on the spatial elements extraction and spatial relation extraction.

\subsection{Experimental Results}
The results of spatial role extraction on SpaceEval is showed in Table \ref{roleresults} and its performance is similar with \citet{shin-etal-2020-bert}. In the stage of CTE, we get 3096 candidate triplets, in which 1355 triplets are positive and 1741 triplets are negative. These figures show that the number of the negative instances is more than that of the positive ones. If we use the null value to construct the candidate triplets, the large number of negative instances will harm the performance critically.

Table \ref{mainresults} shows the overall performance of the baselines and our HMCGR on SpaceEval. Compared with the SOTA baseline BERT-base, our HMCGR significantly improves the overall F1-score by 9.7, especially the Recall with a gain of 19.4. This result verifies the effectiveness of HMCGR, and indicates that our generation model GEN and our classification model CLS can promote each other. Moreover, the improvement comes from all three links QSLINK, OLINK, MOVELINK with the gains of 10.6, 8.4, and 11.8, respectively. This result shows that our HMCGR works well on all links. It is worth noting that our improvement mainly comes from the recall and this indicates that the generation model is helpful to recover those null-role relations.

\begin{table}
\centering
\begin{tabular}{llll}
\hline
Model & P & R & F1\\
\hline
BERT-base& 44.5  &  31.7	& 37.0\\
HMCGR	& 46.7	& 40.0	& 43.0\\
\hline
\end{tabular}
\caption{\label{defaultvalue}
The results of spatial relation extraction on null-role relations.
}
\end{table}

\section{Analysis}

\subsection{Analysis on Null-role Relations}
To further verify the effectiveness of our GEN, we count the null-role relations and Table \ref{defaultvalue} shows the performance of BERT-base and HMCGR. Compared with BERT-base, HMCGR improves the F1-score by 6.0, especially the significant gain on recall (+8.3). This result verifies our motivation that the generation model GEN is effective to extract those null-role relations. However, only 40.0\% of null-role relations in the test set are extracted by GEN and this indicates that the null-role relation extraction has much room for improvement.

\begin{table}
\centering
\begin{tabular}{llll}
\hline
Model & P & R & F1\\
\hline
HMCGR	& \textbf{64.3}	& \textbf{79.2}	& \textbf{70.9}\\
\hline
GEN & 60.4	& 53.1	& 56.5\\
CLS	& 62.0	& 65.5	& 63.7\\
GEN+CLS & 64.1	& 75.2	& 69.2\\
GEN+RFX & 62.2	& 55.1	& 58.8\\
CLS+RFX & 62.0	& 62.5	& 62.2\\
\hline
\end{tabular}
\caption{\label{table:abd}
Ablation study on different modules.
}
\end{table}

\subsection{Ablation Study on Different Modules}
We conduct the ablation experiments to verify the effectiveness of the modules used in HMCGR, and Table \ref{table:abd} shows the results of the simplified models.

The performance descents of both single GEN and CLS are very large, in comparison with the hybrid HMCGR. This result indicates a single classification or generation model maybe cannot extract those null-role and non-null-role relations simultaneously.
Moreover, the performance of GEN is lower than that of CLS and the reason is that the number of non-null-role relations is twice as much as that of null-role relations. Besides, CLS works better than BERT-base and this verifies the success of our classification model. However, the performance of GEN is lower than that of BERT-base and this indicates how to apply generation model to the traditional classification tasks still is a challenge.

The combination model GEN+CLS outperforms GEN and CLS, with huge gains of 12.7 and 5.5, respectively. This indicates GEN  and CLS can boost each other to improve THE F1-score, especially the recall.  In the SpaceEval dataset, 32.3\% of the spatial relations belong to null-role one, in which 65.3\% of the null-role relations do not have the role $trigger$ and the others do not have the role $landmark/goal$. Our GEN can recover almost 40\% null-role relations and this indicates that the generation model prefers to extract those null-role relations. Moreover, the decision from two different models also can further improve the performance on different perspectives.

Compared with GEN, GEN+RFX improves the F1-score by 2.3 with the gains in both the precision and recall. This indicates that our reflexivity evaluation mechanism RFX can not only help the generation model to extract more spatial relations, but also filter out the pseudo relations. However, the F1-score of CLS+RFX is lower than that of CLS, especially the recall. Among three relation types, only the F1-score of MOVELINK decreases from 62.5 to 61.0. The reason is that some triggers in MOVELINKs do not have an antonym (e.g., "run" and "biking") and some sentences cannot be inverted. Besides, compared with GEN+CLS, HMCGR improves the F1-score by 1.7, with a gain of 4.0 on the recall. This verifies that RFX is helpful to discover more relations in a hybrid model.

\begin{table}
\centering
\begin{tabular}{llll}
\hline
Model & P & R & F1\\
\hline
HMCGR	& 64.3	& 79.2	& 70.9\\
\hline
w/o GCN	& 63.3	& 74.7	& 68.1\\
w/o CrossAtt & 62.1	& 74.2	& 67.6\\
\hline
\end{tabular}
\caption{\label{table-cls}
Results of HMCGR and its simplified version on SpaceEval.
}
\end{table}

\subsection{Analysis on CLS}
To verify the contributions of the components in CLS, We conduct the following two simplified versions of HMCGR: 1) w/o GCN: the GCN layer is removed from HMCGR; 2) w/o CrossAtt: the cross attention is removed. That is, we only use BERT to encode sentences. 

Table \ref{table-cls} shows the results of HMCGR and its simplified versions. If we remove the GCN layer and the cross attention, the F1-score will decrease by 2.9 and 3.3, respectively. This result indicates that T5 is helpful for BERT to represent the sentence from different perspectives. As for the GCN layer, we find out that the coreference edge is the main contributor, and more than 90\% of the improvement comes from this edge type.

\subsection{Error Analysis}
The errors of our HMCGR mainly come from those in CTE, GEN, and entity coreference. In table \ref{roleresults}, we can find out that 8.8\% of spatial relations are missing and 11.9\% of pseudo relations are introduced to the following modules by CTE. 

Our statistics on the results shows that GEN often badly predicts those null-role relations when there are a non-null-role relation  and a null-role relation in the same sentence. Since T5 is a sequential generation model, the generation of the next spatial relation will be affected by the relation predicted above. That is, if the previous relation is non-null-role one, the current relation tends to be non-null-role too. Take Table \ref{exa} as an example, there are two MOVELINKs in the sentence. After HMCGR has extracted the first relation MOVELINK(cattle, to, fields), it tends to predict the next one as MOVELINK(men, to, fields), instead of MOVELINK(cattle, null, fields).

Although the coreference edge is the most effective one in the graph, lots of errors derive from it  due to its low performance.

\begin{table}
\centering
\begin{tabular}{l}
\hline 
\textbf{Sentence:} There were already old men taking cattle \\out to the fields to graze.\\
\hline 
\textbf{Gold MOVELINKs:} \\
\{$mover$: cattle, $trigger$: to, $goal$: fields\}\\
\{$mover$: men, $trigger$: null, $goal$: fields\}\\
\hline 
\textbf{Predicted MOVELINKs:} \\ 
\{$mover$: cattle, $trigger$: to, $goal$: fields\}\\
\{$mover$: men, $trigger$: to, $goal$: fields\}\\
\hline 
\end{tabular}
\caption{\label{exa}
Examples of the errors in GEN.
}
\end{table}

\section{Conclusion}
In this paper, we propose a novel hybrid model HMCGR for spatial relation extraction. The generation model GEN can generate those null-role relations, while the classification model CLS can extract those non-null-role relations to complement each other. Moreover, a reflexivity evaluation mechanism is applied to further improve the accuracy based on the reflexivity of spatial relation. Experimental results on the SpaceEval dataset show that our HMCGR outperforms the SOTA baseline significantly. Our future work will focus on how to extract those null-role relations effectively. 

\bibliography{anthology,acl_latex}

\appendix
\end{document}